\documentclass[journal]{IEEEtran}

\usepackage{microtype}
\usepackage{subfigure}
\usepackage{booktabs} 

\usepackage{hyperref,enumitem}

\usepackage{amsmath}
\usepackage{amsfonts}
\usepackage{amssymb}
\usepackage{graphicx,xcolor}
\usepackage{amsfonts}
\usepackage{epstopdf}
\usepackage{algorithm}
\usepackage{algorithmic}
\usepackage{amsmath,amssymb,amsthm}
\usepackage{url}
\usepackage{dsfont}
\usepackage{float}
\ifpdf
  \DeclareGraphicsExtensions{.eps,.pdf,.png,.jpg}
\else
  \DeclareGraphicsExtensions{.eps}
\fi

\newtheorem{theorem}{Theorem}
\newtheorem{proposition}{Proposition}
\newtheorem{corollary}[theorem]{Corollary}

\newtheorem{result}{Result}

\newtheorem{example}{Example}
\newtheorem*{notation}{Notation}
\usepackage{amsopn}

\DeclareMathOperator*{\argmax}{arg\,max}
\DeclareMathOperator*{\argmin}{arg\,min}
\DeclareMathOperator*{\argsup}{arg\,sup}

\newcommand{\Te}{\theta}
\newcommand{\expcst}{J}
\newcommand{\statespace}{\mathcal{X}}
\newcommand{\actionspace}{\mathcal{A}}
\newcommand{\trans}{\mathcal{T}}
\newcommand{\disc}{\gamma}
\newcommand{\reward}{r}
\newcommand{\state}{x}
\newcommand{\tim}{k}
\newcommand{\action}{a}
\newcommand{\real}{\mathbb{R}}
\newcommand{\stdim}{p}
\newcommand{\actdim}{q}

\newcommand{\policy}{\mu}
\newcommand{\bpolicy}{\boldsymbol{\mu}}
\newcommand{\probmeas}{\mathbb{P}}
\newcommand{\inducedPM}{\mathbb{P}^{x_0}_{\boldsymbol{\mu}}}
\newcommand{\inducedEM}{\mathbb{E}^{x_0}_{\boldsymbol{\mu}}}
\newcommand{\inducedEMT}{\mathbb{E}^{x_0}_{\mu_{\theta}}}
\newcommand{\Ulim}{N}

\newcommand{\Blc}{\Big\{}
\newcommand{\Brc}{\Big\}}
\newcommand{\Brn}{\Big)}
\newcommand{\Bln}{\Big(}
\newcommand{\Mu}{\Sigma}
\newcommand{\AlnN}[1]{\begin{align*}#1\end{align*}}
\newcommand{\Aln}[1]{\begin{align}#1\end{align}}

\newcommand{\Thmw}[1]{\begin{theorem}#1\end{theorem}}

\newcommand{\Prf}[2][default]{\begin{proof}(#1) #2\end{proof}}
\newcommand{\Prop}[1]{\begin{proposition} #1 \end{proposition}}
\newcommand{\Corr}[1]{\begin{corollary} #1 \end{corollary}}
\newcommand{\EE}{\mathbb{E}}
\newcommand{\given}{|}
\newcommand{\Qu}{Q}
\newcommand{\ergod}{\pi}
\newcommand{\Bls}{\Big[}
\newcommand{\Brs}{\Big]}
\newcommand{\Rem}{\underline{\textit{Remark}}:~}
\newcommand{\Res}[2][default]{\begin{result} #1 \emph{#2}\end{result}}
\newcommand{\Dis}{\underline{\textit{Discussion}}:~}
\newcommand{\Exmp}[1]{\begin{example} \emph{ #1 }\end{example}}

\newcommand{\defeq}{\overset{\Delta}{=}}
\newcommand{\simT}{T}
\newcommand{\simtime}{t}
\newcommand{\statesim}{s}
\newcommand{\poT}{\oplus}
\newcommand{\neT}{\ominus}
\newcommand{\anab}{\hat{\nabla}}
\newcommand{\actionsim}{a^s}
\newcommand{\pathcst}{\mathcal{R}}
\newcommand{\Bgiven}{\Big|}
\newcommand{\dimtheta}{d}
\newcommand{\arbindxp}{m}
\newcommand{\arbindxn}{n}
\newcommand{\hatpol}{\hat{\policy}}

\newcommand{\randsamp}{z}
\newcommand{\inducedPMP}{\mathbb{P}^{\state_0}_{\policy_{\Te}^{\oplus}}}
\newcommand{\inducedPMN}{\mathbb{P}^{\state_0}_{\policy_{\Te}^{\ominus}}}
\newcommand{\inducedEMP}{\mathbb{E}^{\state_0}_{\policy_{\Te}^{\oplus}}}
\newcommand{\inducedEMN}{\mathbb{E}^{\state_0}_{\policy_{\Te}^{\ominus}}}

\newcommand{\simspace}{\mathcal{S}}
\newcommand{\normal}{\mathcal{N}}
\newcommand{\tpose}{\prime}
\newcommand{\feature}{\phi}
\newcommand{\lip}{L}
\newcommand{\nrm}{\|}
\newcommand{\CapT}{\Theta}
\newcommand{\noise}{w}
\newcommand{\filtr}{\mathcal{F}}
\newcommand{\const}{Y}
\newcommand{\y}{Y}
\newcommand{\w}{W}
\newcommand{\z}{Z}
\newcommand{\fullint}{I}
\newcommand{\negFI}{B}
\newcommand{\nat}{\mathbb{N}}
\newcommand{\indF}{\chi}
\newcommand{\stepsize}{\epsilon}
\newcommand{\taxicab}{d}
\newcommand{\Kanto}{\mathcal{K}}
\newcommand{\Bigtim}{K}
\newcommand{\mDel}{\varepsilon}
\newcommand{\dtim}{d}
\newcommand{\BigO}{O}
\newcommand{\SamCmp}{M^{\Delta}_{\disc}}
\newcommand{\var}{\text{Var}}
\newcommand{\varWD}{G_{WD}}
\newcommand{\varSF}{G_{SF}}
\newcommand{\extra}{\eta}
\newcommand{\shifttim}{\bar{\tim}}
\newcommand{\Bigabs}{\Big|}
\newcommand{\Upbound}{M}
\newcommand{\cst}{c}
\newcommand{\act}{u}

\title{Policy Gradient using Weak Derivatives \\ for Reinforcement Learning}
\author{Sujay Bhatt, Alec Koppel, Vikram Krishnamurthy
\thanks{Sujay Bhatt and Vikram Krishnamurthy are with the Dept. of Electrical and Computer Engineering, Cornell University (\{sh2376, vikramk\}@cornell.edu).  Alec Koppel is with U.S. Army Research Laboratory  (alec.e.koppel.civ@mail.mil).}
}

\begin{document}
\maketitle



\begin{abstract}
This paper considers policy search in continuous state-action reinforcement learning problems. Typically, one computes search directions using a classic expression for the policy gradient called the Policy Gradient Theorem, which decomposes the gradient of the value function into two factors: \textit{the score function} and the $Q-$function. This paper presents four results: (i)~an alternative policy gradient theorem using weak (measure-valued) derivatives instead of score-function is established; (ii)~the stochastic gradient estimates thus derived are shown to be unbiased and to yield algorithms that converge almost surely to stationary points of the non-convex value function of the reinforcement learning problem; (iii)~the sample complexity of the algorithm is derived and is shown to be $O(1/\sqrt{k})$; (iv)~finally, the expected variance of the gradient estimates obtained using weak derivatives is shown to be lower than those obtained using the popular \textit{score-function} approach. Experiments on OpenAI gym pendulum environment show superior performance of the proposed algorithm.
\end{abstract}
\section{Introduction}
\label{sec:intro}

Reinforcement Learning (RL) is a form of implicit stochastic adaptive control where the optimal control policy is estimated without directly estimating the underlying model. This paper considers reinforcement learning for an infinite horizon discounted cost continuous state Markov decision process. In a MDP, actions affect the Markovian state dynamics and result in rewards for the agent. The objective is to find a map from the states to actions, also known as policy, that results in the accumulation of largest expected return~\cite{Bellman:1957}. There are many approaches to estimate a policy: policy iteration, $Q-$learning \cite{Watkins1989,kqlearning} (which operates in ``value" space \cite{pkgtd}), policy-gradients \cite{SMSM00,policy_grad} (that operate in policy space); see~\cite{Ber05,SB17}.

Recently, policy-gradient algorithms have gained popularity due to their ability to address complex real-world RL problems with \textit{continuous state-action spaces}. Given a parametrized policy space, usually designed to incorporate domain knowledge, policy-gradient algorithms update policy parameters along an estimated ascent direction of the expected return. Depending on whether the expected reward or the value function is convex or non-convex, the parameters converge to a minimum or a stationary point; for a comprehensive survey see~\cite{GBLB12,DNP13}.

Typically, to compute the ascent direction in policy search~\cite{Sil09}, one employs the Policy Gradient Theorem~\cite{SB17} to write the gradient as the product of two factors: the $Q-$function{\footnote{$Q-$function is also known as the state-action value function \cite{SB17}. It gives the expected return for a choice of action in a given state.}} and the score function (a likelihood ratio). This score function approach has yielded numerous viable policy search techniques~\cite{Wil92,Doy00,GBB04,SB17}, although the resulting  gradient estimates are afflicted with high variance: the score function is a martingale and so for a Markov process its variance is $O(N)$ for $N$ measurements. In pursuit of reducing the variance, we propose replacing  the score function with the Jordan decomposition of signed measures~\cite{Bil08}, similar to the method{\footnote{Jordan decomposition (also known as Hahn-Jordan decomposition) of signed measures is a specific type of weak derivative form - this expresses the derivative of a measure as the weighted difference of orthogonal measures. For example, the gradient of gaussian  policy~\cite{Doy00} is written as a (scaled) difference of two Rayleigh  policies.}} of \textit{Weak Derivatives} in the finite state-action MDP literature; see \cite{glasserman1991gradient,KA12,Kri16}. 

To estimate the $Q-$function in the policy gradient~\cite{SB17}, we use 
Monte Carlo roll-outs with random path lengths akin to \cite{Pat18}, motivated by the fact that obtaining unbiased estimates of continuous state-action $Q-$function in the infinite horizon case is otherwise challenging. The product of these terms yields a valid estimate of the overall policy gradient, as in \cite{SB17}. 

This paper considers reinforcement learning for the case when the underlying system can be simulated using statistically independent trials with different policies. Our main results are:
\begin{enumerate}
\item A policy gradient theorem using Jordan decomposition for the policy gradient. We establish that the resulting policy gradient algorithm, named Policy Gradient with Jordan Decomposition (PG-JD), yields unbiased estimates of the gradient of the reward function.
\item to establish that the PG-JD algorithm converges to a stationary point of the parametrized value function almost surely under decreasing step-sizes.
\item to derive the iteration (and sample{\footnote{Iteration complexity is a measure of the number of changes of the unknown parameter. Sample complexity includes the additional simulations required to estimate the continuous state-action $Q-$function using Monte Carlo roll-out with random path lengths.}}) complexity as $O(1/ \sqrt{k})$, where $k$ is the time step. This shows that the convergence rate is similar to stochastic gradient method for non-convex settings.
\item to upper-bound the expected variance of the gradient estimates obtained using the PG-JD algorithm, which isshown to be lower than those generated by score function methods using Monte Carlo roll-outs with random path lengths, for common policy parametrizations.
\end{enumerate}
The setup and problem formulation are discussed in Sec.~\ref{sec:PF}. The new policy gradient theorem using weak derivatives (Jordan decomposition) is derived in Sec.~\ref{sec:PGJD}. The algorithm to compute the stochastic gradient and the policy parameter update is given in Sec.~\ref{sec:SGJD}. Convergence analysis of the stochastic gradient ascent algorithm and its statistical properties are derived in Sec.~\ref{sec:Prop_Alg}. Numerical studies on OpenAI gym using the pendulum environment is discussed in Sec.~\ref{sec:NumStu}.



\section{Problem Formulation and Policy Search} \label{sec:PF}
The problem of reinforcement learning is considered in the framework of Markov Decision Process, which is defined as a tuple $(\statespace,\actionspace,\trans,\reward,\disc)$ consisting of the \textit{state space} $\statespace \subseteq {\real}^{\stdim}$, a subset of Euclidean space with elements $\state \in \statespace$; the \textit{action space} $\actionspace \subseteq {\real}^{\actdim}$, a subset of Euclidean space with elements $\action \in \actionspace$; the \textit{transition law} $\trans$, a probability density function $\trans(\cdot \given \action,\state) \in \probmeas(\statespace)$ that assigns a next-state upon taking action $\action$ in state $\state$, where $\probmeas(\statespace)$ denotes the set of all probability measures on $\statespace$; the \textit{reward function} $\reward (\state,\action)$, a real valued function on the product space $\statespace \times \actionspace$; the \textit{discount} $\disc \in (0,1)$, a parameter that scales the importance of future rewards.

A \textit{stochastic Markov policy} $\bpolicy = \{\policy_\tim\}$ is defined as a sequence of transition probabilities from $\statespace$ to $\actionspace$ such that $\policy_\tim (D(\state) \given \state) = 1$ for each $\state \in \statespace$ and $\tim = 0,1,\cdots$. Here $D$ maps  each $\state \in \statespace$ to the set of all available actions $D(x)$. Let $\Mu$ denote the class of stochastic Markov policies.

For an initial state $\state_0$ and a stochastic Markov policy $\bpolicy \in \Mu$, define the expected reward function
\begin{equation} \label{eq:ValFu}
\expcst(\state_0,\bpolicy) = \lim_{\Ulim \rightarrow \infty} \inducedEM \Blc \sum_{\tim=0}^{\Ulim} \disc^\tim \reward(\state_\tim,\action_\tim)~ \Bgiven \action_\tim \sim \policy_{\tim}(\cdot \given \state_\tim)\Brc
\end{equation}
For an initial state $\state_0$ and a Markov policy $\bpolicy \in \Mu$, using Ionescu Tulcea theorem \cite{Nev65,HL12}, define $\inducedPM$ as 
\begin{align}\label{eq:ITthm}
\inducedPM(d\state_0 d\action_0 \cdots d\state_\tim d\action_{\tim} \cdots) &= \policy_0(d\state_0) \prod_{\tim=1}^{\infty} \policy_{\tim} (d\action_\tim \given \state_{\tim})  \\  &\quad \times\trans(d\state_{\tim} \given \state_\tim, \action_\tim).\nonumber
\end{align}
%
Here $\policy_0 \in \probmeas(\statespace)$ is an atomic measure with $\policy_0(\state_0) = 1$. The expectation $\inducedEM$ in~(\ref{eq:ValFu}) is with respect to $\inducedPM$ in~(\ref{eq:ITthm}).
Our goal is to find the policy $\bpolicy$ that maximizes the long-term reward accumulation, or value:
\begin{align} \label{eq:MP}
\!\!\bpolicy^* \!\!= \argsup_{\bpolicy \in \Mu}  \lim_{\Ulim \rightarrow \infty} \inducedEM \Blc \sum_{\tim=0}^{\Ulim} \!\! \disc^\tim \reward(\state_\tim,\action_\tim) \Bgiven \!\! ~ \action_\tim \! \sim \! \policy_\tim(\cdot \given \state_\tim)\!\Brc.
\end{align}
For the infinite horizon problem~(\ref{eq:MP}), it is sufficient \cite{Bla65,BS78,HL12,Fei96} to restrict the class $\Mu$ of policies to the class $\Mu_s \subset \Mu$ of stationary stochastic Markov policies. A \textit{stationary stochastic Markov policy} $\bpolicy (= \{\policy\}) \in \Mu_s$ is defined as the transition probability from $\statespace$ to $\actionspace$ such that $\policy (D(\state) \given \state) = 1$ for each $\state \in \statespace$. 
In order to solve (\ref{eq:MP}) we resort to direct policy search over the space of continuous stationary  policies. It is convenient to \textit{parametrize} the stationary policy $\policy(\cdot \given \cdot)$ as $\policy_\Te (\cdot \given \cdot)$ for $\Te \in \CapT \subseteq \real^{\dimtheta},$ for $\dimtheta \in \mathbb{N}$, and search over the space of $\Te$. For example, consider Gaussian policy $\policy_{\Te}(\cdot \given \state) = \normal(\Te^{\tpose}\feature(\state), \sigma^2)$. Here the function $\feature(\cdot)$ is commonly referred to as the feature map and $\sigma$ denotes the standard deviation. With a slight abuse of notation, the problem (\ref{eq:MP}) can be reformulated in terms of the finding a parameter vector $\Te$ to satisfy:
\begin{align} \label{eq:MP_p}
\Te^* &= \argmax_{\Te \in \real^{\dimtheta}} \expcst (\Te),\\
\expcst(\Te) &=  \lim_{\Ulim \rightarrow \infty} \inducedEMT \Blc \sum_{\tim=0}^{\Ulim} \disc^\tim \reward(\state_\tim,\action_\tim)~ \Bgiven \action_\tim \sim \policy_{\Te}(\cdot \given \state_\tim)\Brc. \nonumber 
\end{align}  
Here $\inducedEMT$ is the expectation with respect to the measure induced by the probability measure as in (\ref{eq:ITthm}) with the policy $\bpolicy_{\Te} = \{\policy_{\Te} \}$ and initial state $\state_0$.


\section{Policy Gradient Theorem via Hahn-Jordan} \label{sec:PGJD}
The foundation of any valid policy search technique is a valid ascent direction on the value function with respect to the policy parameters. Classically, one may derive that the policy gradient decomposes into two factors: the action-value (Q) function and the score function \cite{SMSM00}. Here we establish that one may obviate the need for the log trick that gives rise to the score function through measure-valued differentiation by employing the Jordan decomposition of signed measures \cite{Bil08}. To begin doing so, define the $Q-$function as 
\begin{equation}\label{eq:Q_function}
\Qu_{\policy_{\Te}}(\state,\action) = \EE_{\policy_{\Te}} \Blc \sum_{\tim=0}^{\infty} \disc^{\tim} r(\state_{\tim},\action_{\tim}) \Bgiven \state_0 = \state, \action_0 = \action \Brc \; .%
\end{equation}
 The weak derivative of the signed measure $\nabla \policy_{\Te}(\cdot \given \state)$ using Jordan decomposition
{ \parindent=0pt\footnote{\vspace{-4mm} \Res[\cite{Bil08}]{[Hahn Decomposition] Let $\mu$ be a finite signed measure on the measurable space $(\Omega,\mathcal{F})$. There exists a disjoint partition of the set $\Omega$ into $\Omega^+$ and $\Omega^-$ such that $\Omega = \Omega^+ \cup \Omega^-$, $\mu(A) \geq 0, \forall A \subset \Omega^+$, and $\mu(B) \leq 0, \forall B \subset \Omega^-$.  
}
\Res[\cite{Bil08}]{[Jordan Decomposition]
Every finite signed measure $\mu$ has a unique decomposition into a difference $\mu = \mu^+ - \mu^-$ of two finite non-negative measures $\mu^+$ and $\mu^-$ such that for any Hahn decomposition $(\Omega^+,\Omega^-)$ of $\mu$, we have for $A \in \mathcal{F}$ that $\mu^+(A) = 0$ if $A \subset \Omega^-$ and $\mu^-(A) = 0$ if $A \subset \Omega^+$.
}}} 
is given as 
\begin{equation}\label{eq:jordan_decomposition}
\nabla \policy_{\Te}(\cdot \given \state) = g(\Te,\state) \Blc \policy^{\poT}_{\Te}(\cdot \given \state) - \policy^{\neT}_{\Te}(\cdot \given \state) \Brc
\end{equation}
Here the decomposed positive and negative component measures  $\policy^{\poT}_{\Te}(\cdot \given \state)$ and $\policy^{\neT}_{\Te}(\cdot \given \state)$ are orthogonal in $L^2$ (see Example~\ref{ex:GEP} below). The ergodic measure associated with the transition kernel $\trans(\cdot \given \state_0,\action_0)$ and policy $\policy_{\Te}$ is $\ergod_{\policy_\Te}(\state) = (1-\disc)\sum_{\tim=0}^\infty \disc^\tim \cdot \trans(\state_\tim = \state \given \state_0,\policy_\Te)$. The induced measures on $\statespace \times \actionspace$ by $\policy^{\poT}_{\Te}$ and $ \policy^{\neT}_{\Te}$ are defined as $\policy^{\poT}_{\Te} (\state,\action) \defeq \policy^{\poT}_{\Te}(\action \given \state) \cdot \ergod_{\policy_\Te}(\state)$ and $\policy^{\neT}_{\Te} (\state,\action) \defeq \policy^{\neT}_{\Te}(\action \given \state) \cdot \ergod_{\policy_\Te}(\state)$. Using this measure (weak) derivative representation of the policy, we can write the gradient of the value function with respect to policy parameters $\theta$ in an unusual way which is given in the following theorem.
%
\Thmw{\label{thm:PGJD}(Jordan Decomposition for Policy Gradients) The policy gradient using Jordan decomposition takes the form
\Aln{\label{eq:policy_gra}
\nabla \expcst(\theta) &= \frac{1}{1-\disc} \Bls \EE_{(\state,\action)\sim \policy^{\poT}_{\Te}(\cdot,\cdot)} \Blc g(\Te,\state) \cdot \Qu_{\policy_\Te}(\state,\action) \Brc  \\
& \qquad\qquad-  \EE_{(\state,\action)\sim \policy^{\neT}_{\Te}(\cdot,\cdot)} \Blc g(\Te,\state) \cdot \Qu_{\policy_\Te}(\state,\action) \Brc \Brs.\nonumber } 
where $g(\Te,\state)$ is a normalizing constant to ensure $\policy^{\poT}$ and $\policy^{\neT}$ are valid measures.}
\Dis{
A proof is included in the Appendix. Theorem~\ref{thm:PGJD} is the policy gradient theorem using weak derivatives, specifically Jordan decomposition. In Theorem~\ref{thm:PGJD}, note that the $\Qu$ functions in the expectations are the same, indicating that the model is unaffected by the measure decomposition; only the induced measures are different. The expression for the gradient in~(\ref{eq:policy_gra}) contains a difference of two expectations. Unlike, the method of score functions, the expectation obviates the need for a \emph{score function} term. Intuitively, this allows us to avoid computing the logarithm of the policy which may amplify useless parts of the state-action space and cause variance to needlessly be increased, and instead yield a sharp ``perceptron-like" behavior. In subsequent sections, we indeed establish that this representation may reduce variance but this reduction intrinsically depends on the policy parameterization. Note that $g(\Te,\state)$ for a given parameter $\Te$ and state $\state$, is a constant, which makes the stochastic gradient easier to compute in Algorithm~\ref{alg:SGJD}. Before continuing, we present a representative example.
}
\Exmp{ \label{ex:GEP}
Consider a gaussian  policy $\policy_{\Te}(\cdot \given \state) = \normal(\Te^{\tpose}\feature(\state), \sigma^2)$, where the mean of the gaussian distribution is modulated by the optimization parameter. The Jordan decomposition of the gaussian  policy can be derived as follows:
\Aln{
\policy_{\Te}(\cdot \given \state) &= \normal(\Te^{\tpose}\feature(\state), \sigma^2) = \frac{1}{\sqrt{2\pi \sigma^2}} \exp \Bln\frac{(a-\Te^{\tpose}\feature(\state))^2}{2\sigma^2}\Brn. \\
\nabla \policy_{\Te}(\cdot \given \state) &=  \frac{1}{\sqrt{2\pi \sigma^2}} \exp \Bln\frac{(a-\Te^{\tpose}\feature(\state))^2}{2\sigma^2}\Brn \nonumber \\
 &\quad \times\frac{1}{\sigma^2}(a-\Te^{\tpose}\feature(\state)) \cdot \feature(\state). \nonumber \\
%
&:= g(\Te,\state) \Blc \policy_{\Te}^{\poT}(\cdot \given \state) - \policy_{\Te}^{\neT}(\cdot \given \state)\Brc, 
}
Here we may glean the normalizing constant $g(\Te,\state) = \frac{\feature(\state)}{\sqrt{2\pi \sigma^2}}$ and the positive and negative component measures are
\Aln{\policy_{\Te}^{\poT}(\cdot \given \state) &= \frac{1}{\sigma^2}(a-\Te^{\tpose}\feature(\state)) \cdot \exp \Bln\frac{(a-\Te^{\tpose}\feature(\state))^2}{2\sigma^2}\Brn, \\
\policy_{\Te}^{\neT}(\cdot \given \state) &= \frac{1}{\sigma^2}(\Te^{\tpose}\feature(\state)-a) \cdot \exp \Bln\frac{(a-\Te^{\tpose}\feature(\state))^2}{2\sigma^2}\Brn. 
}
Observe that $\policy_{\Te}^{\poT}(\cdot \given \state)$ and $\policy_{\Te}^{\neT}(\cdot \given \state)$ define the Rayleigh{\footnote{The probability density function corresponding to Rayleigh distribution is: $f(\state) = \frac{\state}{\sigma^2} \cdot \exp \Bln \frac{\state^2}{2 \sigma^2}\Brn$, $\state \geq0$.}}  policy. They are orthogonal in the sense that $\policy_{\Te}^{\poT}(\cdot \given \state)$ is defined on {\footnote{$\indF(\cdot)$ denotes the indicator function.}} $\indF(a > \Te^{\tpose}\feature(\state))$ and $\policy_{\Te}^{\neT}(\cdot \given \state)$ is defined over $\indF(a < \Te^{\tpose}\feature(\state))$. 
}

%
\section{Policy Search via Jordan Decomposition} \label{sec:SGJD}

In order to develop a policy search method based on Theorem~\ref{thm:PGJD}, we need samples of both factors inside the expectation in~\eqref{eq:policy_gra} which are unbiased. We first focus on the later factor, the $Q-$function.
\subsection{Estimating the Action-Value}\label{sec:estQ}
The estimation of the $Q-$function is carried out using Monte Carlo roll-outs of random path lengths, similar to \cite{Pat18}. Here the random length is a geometric random variable with parameter $\disc$, the discount factor in the reinforcement learning problem. Specifically, we simulate $T\sim\text{ Geom }(1-\gamma)$ and then simulate state-action pairs according to the positive and negative induced policies $\pi^\poT$ and $\pi^\neT$. For this time horizon, we collect rewards for the two different trajectories.

More specifically, from a given starting state $\state_0$, a (real) trajectory is simulated to update the policy parameters $\Te$. At each epoch $\tim$ of the parameter update $\Te_{\tim}$, the simulator (modeled as $(\simspace (=\statespace),\actionspace,\trans,\reward,\disc)$) is called two times to simulate two different (phantom{\footnote{Here the word ``phantom" is used to refer to the actions on the simulator.}}) trajectories. These trajectories correspond to the random Monte-Carlo roll-outs used to estimate the $Q-$functions with two different policies, the positive and negative policy measure, and hence the stochastic gradient of the expected reward function. Let $\simT$ denote a geometrically distributed random variable: $\simT \sim \text{Geom}(1-\disc)$ where $\disc$ is the discount factor. Let the path-wise cost be defined by $\pathcst_{\policy_{\theta}}^{\simT} = \sum_{\tim=0}^{\simT} r(\state_{\tim},\action_{\tim}) \Bgiven \action_{\tim} \sim  \policy_{\Te}(\cdot \given \state_{\tim})$.

\Dis{
Algorithm~\ref{alg:SGJD} with Algorithm~\ref{alg:EQ} is the stochastic gradient algorithm that is used to update the policy parameters. The simulation consists of a single simulation (real trajectory) to update the parameters and multiple phantom simulations to estimate the gradient of the expected reward function. The two phantom trajectories correspond to different polices and not different models, starting from the system's state represented by the state corresponding to the real trajectory. The stochastic gradient computation is summarized in three steps: For a fixed initial state-- (i) Simulate two phantom initial actions from the measures obtained using Jordan decomposition, i.e, $\policy^{\neT}_{\Te_{\tim}}(\cdot \given \statesim^{\neT}_0)$ and $\policy^{\poT}_{\Te_{\tim}}(\cdot \given \statesim^{\poT}_0)$. (ii) Simulate a geometric random variable $\simT_{\tim}$, and (iii) Perform Monte Carlo roll-outs of length $\simT_{\tim}-1$ (i.e, simulate and feed actions to the simulator and collect the rewards) using the system policy derived from old parameters, i.e using $\{\policy_{\Te_{\tim}}(\cdot \given \statesim^{\poT}_\act) \}_{\act=1}^{\act=\simT_{\tim}-1} \}$ and $\{\policy_{\Te_{\tim}}(\cdot \given \statesim^{\neT}_\act) \}_{\act=1}^{\action=\simT_{\tim}-1} \}$.}

The merit of using these random horizons for estimation of the Q function, as summarized in Algorithm \ref{alg:EQ}, is that one may  establish that it is an unbiased estimate in the infinite-horizon discounted case, as we summarize in the following theorem. 

\Thmw{\label{thm:UnbEst}
For a geometric r.v $\simT$, let the approximate state-action value function (Q-function) be defined by $\hat{\Qu}_{\policy_{\Te}}(\state,\action; \simT) = \EE_{\policy_{\Te}} \Blc \sum_{\tim=0}^{\simT} r(\state_{\tim},\action_{\tim}) \Bgiven \state_0 = \state, \action_0 = \action  \Brc$.  Let $\simT$ denote a geometrically distributed random variable. Then,
\Aln{
\EE_{\policy_{\Te}} \Blc \pathcst_{\policy_{\Te}}^{\simT} \Brc &= \hat{\Qu}_{\policy_{\Te}}(\state,\action; \simT). \label{eq:PaCs}\\
\EE_{\simT} \Blc \hat{\Qu}_{\policy_{\Te}}(\state,\action; \simT) \Brc &=  \Qu_{\policy_{\Te}}(\state,\action).\label{eq:UnQ}
}}

A proof is included in the Appendix. Now that we may obtain unbiased samples of the action-value function, we shift focus to how to compute the stochastic gradients needed for policy search based on Jordan decomposition (Theorem \ref{thm:PGJD}).


%

\begin{algorithm}[t!]
\caption{Unbiased estimation of $\Qu_{\policy}$}
\label{alg:EQ}
\begin{algorithmic}
\STATE{\textbf{Input}: Trajectory length $\simT_{\tim}$, states $\statesim_0 = \statesim^{\poT}_0,\statesim^{\neT}_0$, phantom actions $\actionsim_0 = \action^{\poT}_0, \action^{\neT}_0$, simulator policies $\policy = \policy_{\Te_{\tim}}^{\poT}, \policy_{\Te_{\tim}}^{\neT}$. }
\STATE{\textbf{Output}: Unbiased Q-function estimates: $\pathcst_{\hatpol_{\Te}^{\poT}}^{\simT_\tim}$ and $\pathcst_{\hatpol_{\Te}^{\neT}}^{\simT_\tim}$.}
\STATE{\hspace{1.4cm} Initialize~$\pathcst_{\policy}^{\simT_\tim} \leftarrow 0.$}
\FORALL{$\policy = \policy_{\Te_{\tim}}^{\poT}, \policy_{\Te_{\tim}}^{\neT}$ and $\simtime = 0,1,2,\cdots,\simT_\tim-1$}
\STATE{\hspace{1cm} $\pathcst_{\policy}^{\simT_\tim} \leftarrow \pathcst_{\policy}^{\simT_\tim} + \reward(\statesim_{\simtime},\actionsim_{\simtime}).$}
\STATE{\hspace{1cm} $\statesim_{\simtime+1} \sim \trans(\cdot \given \statesim_{\simtime},\actionsim_{\simtime})$, $\actionsim_{\simtime+1} \sim \mu(\cdot \given \statesim_{\simtime+1})$.}
\ENDFOR
\end{algorithmic}
\end{algorithm}

\subsection{Stochastic Gradient Algorithm} \label{subsec:SGA}

With the estimation of the action-value function addressed, we now discuss how we can sample the former factor: the signed measure gradients. Specifically, Theorem~\ref{thm:PGJD} can be used to effectively compute the gradient given access to an oracle/simulator that may generate state-action-reward triples. 
%
%
%
%
It is well known that one only needs to compute estimates of the gradient that are unbiased in expectation to ensure convergence of the iterates to a stationary point~\cite{SB17}. This results in a modification of the gradient expression as in REINFORCE algorithm \cite{Wil92,SB17}, which is a stochastic gradient, for computing the optimal policy of the reinforcement learning problem. Let $\EE_{\simT}$ denote the expectation with respect to the geometric distribution.

Using Theorem~\ref{thm:UnbEst} and Fubini's Theorem \cite{Bog07}, the gradient in~(\ref{eq:policy_gra}) can be rewritten to make it implementable on a simulator:
\Aln{
\nabla J(\theta) &= \frac{1}{1-\disc} \Bls \EE_{\simT} \Blc \EE_{(\state,\action)\sim \policy^{\poT}_{\Te}(\cdot,\cdot)} \Blc g(\Te,\state) \cdot \hat{\Qu}_{\policy_{\Te}}(\state,\action; \simT) \Brc \nonumber \\ 
&\quad-  \EE_{(\state,\action)\sim \policy^{\neT}_{\Te}(\cdot,\cdot)} \Blc g(\Te,\state) \cdot \hat{\Qu}_{\policy_{\Te}}(\state,\action; \simT) \Brc \Brc \Brs \label{eq:poliy_conv_grad} 
}
We have from Theorem~\ref{thm:UnbEst} and (\ref{eq:poliy_conv_grad}),
\Aln{
\anab J_{\simT}(\Te) &= \frac{g(\Te,\state_0)}{1-\disc} \Bls  \pathcst_{\hatpol_{\Te}^{\poT}}^{\simT} - \pathcst_{\hatpol_{\Te}^{\neT}}^{\simT}  \Brs \label{eq:SGD-JD_T}\\
\anab J(\Te) &= \frac{g(\Te,\state_0)}{1-\disc} \Bls \pathcst_{\hatpol_{\Te}^{\poT}}^{\simT_{\randsamp}} - \pathcst_{\hatpol_{\Te}^{\neT}}^{\simT_{\randsamp}} \Brs \label{eq:SGD-JD}
}
Here the initial state simulated from the ergodic measure is $\state_0 \sim \ergod_{\policy_{\Te}}(\state)$, and the policies that simulate the two trajectories are: $\hatpol_{\Te}^{\poT} \defeq \{\policy^{\poT}_{\Te}, \{\policy_{\Te} \}_l \}, l = 1,2,\cdots$ and $\hatpol_{\Te}^{\neT} \defeq \{\policy^{\neT}_{\Te}, \{\policy_{\Te} \}_l \}, l = 1,2,\cdots$. Here the initial actions are simulated from the decomposed measures and the parametrized policy is used for the remainder of the trajectory simulation. Here~(\ref{eq:SGD-JD_T}) is the (stochastic) gradient estimate for a random path length $\simT$ and (\ref{eq:SGD-JD}) is the (stochastic) gradient estimate using a realization $\simT_{\randsamp}$. Using the estimates \eqref{eq:SGD-JD} that are computable using Algorithm~\ref{alg:EQ} to estimate the Q function with respect to the signed measures, then, we may write out an iterative stochastic gradient method to optimize $\theta$ with respect to the value function as
\Aln{ \label{eq:theta_update}
\Te_{\tim+1} = \Te_{\tim} + \stepsize_\tim \cdot \anab \expcst(\Te_{\tim}) \;.}
The overall policy search routine is summarized as Algorithm~\ref{alg:SGJD}. Its convergence and variance properties are discussed in the following section.

\begin{algorithm}[t!]
\caption{Policy Gradient with Jordan Decomposition (PG-JD)}
\label{alg:SGJD}
\begin{algorithmic}
\STATE{\textbf{Input}: System state $x_{\tim+1}$, parameter vector $\Te_\tim$, and continuous random policy $\policy_{\theta_{\tim}}$. }
\STATE{\textbf{Output}: Parameter $\Te_{\tim+1}$ and next system input $\action_{\tim+1} \sim \policy_{\theta_{\tim+1}}$.}
\STATE{\textbf{Step 1}. Simulate $\simT_\tim \sim$ $\text{Geom}(1-\disc)$, i.e., $P(\simT_\tim= \simtime)=(1-\disc)\disc^{\simtime}$.}
\STATE{\hspace{0.5cm} Define the initial conditions: $\statesim^{\poT}_0, \statesim^{\neT}_0 = \state_{\tim+1}$.}
\STATE{\hspace{0.5cm} Define: $\hatpol_{\Te_{\tim}}^{\poT} \defeq \{\policy^{\poT}_{\Te_{\tim}}(\cdot \given \statesim^{\poT}_0), \{\policy_{\Te_{\tim}}(\cdot \given \statesim^{\poT}_\action) \}_{\action=1}^{\action=\simT_{\tim}-1} \}$ as the policy for trajectory~$1$.}
\STATE{\hspace{0.5cm} Define: $\hatpol_{\Te_{\tim}}^{\neT} \defeq \{\policy^{\neT}_{\Te_{\tim}}(\cdot \given \statesim^{\neT}_0), \{\policy_{\Te_{\tim}}(\cdot \given \statesim^{\neT}_\action) \}_{\action=1}^{\action=\simT_{\tim}-1} \}$ as the policy for trajectory~$2$.}
\STATE{\textbf{Step 2}. Simulate $\action^{\poT}_0 \sim \policy^{\poT}_{\Te_{\tim}}(\cdot \given \statesim^{\poT}_0)$ and $\action^{\neT}_0 \sim \policy^{\neT}_{\Te_{\tim}}(\cdot \given \statesim^{\neT}_0)$.}
\STATE{\textbf{Step 3}. Compute $\Qu_{\hatpol_{\Te_{\tim}}^{\poT}}(\statesim^{\poT}_0,\action^{\poT}_0)$ and $\Qu_{\hatpol_{\Te_{\tim}}^{\neT}}(\statesim^{\neT}_0,\action^{\neT}_0)$ 
using Algorithm~\ref{alg:EQ}.}
\STATE{\textbf{Step 4}. Compute $\anab \expcst(\Te_{\tim}) = \frac{g(\Te_{\tim},\state_{\tim+1})}{1-\disc} \cdot \Blc \pathcst_{\hatpol_{\Te_{\tim}}^{\poT}}^{\simT_{\tim}} - \pathcst_{\hatpol_{\Te_{\tim}}^{\neT}}^{\simT_{\tim}} \Brc$}
\STATE{\textbf{Step 5}. Compute $\Te_{\tim+1} = \Te_{\tim} + \stepsize_\tim \cdot \anab \expcst(\Te_{\tim})$.}
\end{algorithmic}
\end{algorithm}

\section{Convergence, Complexity, $\&$ Variance Analysis} \label{sec:Prop_Alg}
In this section, we discuss a few properties of the stochastic gradient ascent algorithm derived using weak derivatives, namely, convergence, the iteration complexity, sample complexity, and the variance of the resulting gradient estimates. 
\subsection{Convergence Analysis}
We now analyze the convergence of the PG-JD algorithm (Algorithm~\ref{alg:SGJD} ), establishing that the stochastic gradient estimates obtained from the algorithm are unbiased estimates of the true gradient, and that the parameter sequence~\eqref{eq:theta_update} converges almost surely to a stationary point of the value function~\eqref{eq:MP_p}.  To do so, some assumptions are required which we state next.
%
\subsubsection{Assumptions} \label{subse:MoDAs}
\begin{enumerate}[label=(\roman*)]
	\item \label{as:reward} The reward function{\footnote{Let the product space $\statespace \times \actionspace$ be equipped with the taxi-cab norm: 
			\AlnN{
				\taxicab_{\statespace \actionspace}((\state_1,\action_1), (\state_2,\action_2)) &= \taxicab_{\statespace}(\state_1,\state_2) + \taxicab_{\actionspace}(\action_1,\action_2) \nonumber \\ & \forall(\state_1,\state_2,\action_1,\action_2) \in \statespace^2 \times \actionspace^2,}
			where $\taxicab_{(\cdot)}$ denotes the corresponding metric on the Euclidean space.
	}}
	 $\reward(\state,\action)$ is bounded Lipschitz, i.e,
	\AlnN{
		& |\reward(\state,\action)| \leq M (< \infty),~\forall (\state,\action) \in \statespace \times\actionspace. \nonumber \\
		&\forall (\state_1,\state_2,\action_1,\action_2) \in \statespace^2 \times \actionspace^2, \nonumber\\ 
		&| \reward(\state_1,\action_1) - \reward(\state_2,\action_2)| \leq \lip_{\reward} \cdot \taxicab_{\statespace \actionspace}((\state_1,\action_1), (\state_2,\action_2)).
	}
	\item  \label{as:transition_smooth} The transition law{\footnote{As in \cite{Hin05}, $\Kanto(\upsilon,\nu)$ denotes the Kantorovich distance between probability distributions $\upsilon$ and $\nu$. It is given by:
			\AlnN{
				\Kanto(\upsilon,\nu) \defeq \sup_{f} \Blc \Big| \int f d\upsilon - \int f d\nu \Big| : \nrm f \nrm_1 \leq 1 \Brc.
			}
	}} $\trans(\cdot \given \state,\action)$ is Lipschitz, i.e,
	\AlnN{
		\forall (\state_1,\state_2,\action_1,\action_2) \in \statespace^2 \times \actionspace^2, &\nonumber \\ 
		\Kanto \Bln \trans(\cdot \given \state_1,\action_1), \trans(\cdot \given \state_2,\action_2) \Brn \leq \lip_{\trans} \cdot &\taxicab_{\statespace \actionspace}((\state_1,\action_1), (\state_2,\action_2)).
	}
	\item \label{as:transition_ergodic} For $\Te \in \real^{\dimtheta}$, the transition law $\trans(\cdot \given \state, \policy_{\Te})$ is $\psi-$irreducible, positive Harris recurrent, and geometrically ergodic.  
	\item \label{as:policy_smooth} The continuous policy $\policy_{\Te}(\action \given \state)$ is Lipschitz, i.e,
	\AlnN{
		&\forall (\state_1,\state_2) \in \statespace^2, \Te \in \CapT, \\ 
		\Kanto \Bln \policy_{\Te}(\cdot \given \state_1), \policy_{\Te}(\cdot \given \state_2) \Brn &\leq \lip_{\Te} \cdot \taxicab_{\statespace}(\state_1, \state_2).
	}
	\item \label{as:stepsize}
	 $\sum_{\tim} \stepsize_{\tim} = \infty$ and $\sum_{\tim} \epsilon^{2}_{\tim} < \infty$.
	\item \label{as:variance_growth} The stochastic gradient 
	\AlnN{
		\EE \Blc \nrm \anab \expcst(\Te) \nrm^{2} \Brc \leq \arbindxp + \arbindxn \nrm \nabla \expcst(\Te) \nrm^{2}} for all $\Te \in \CapT$, and $\arbindxn,\arbindxp > 0$.
\end{enumerate}
Assumptions \ref{as:reward} - \ref{as:transition_ergodic} are model assumptions, whereas Assumptions \ref{as:policy_smooth} - \ref{as:variance_growth} impose restrictions on how the algorithm behaves. Assumption \ref{as:reward} is standard, and tied to learnability of the problem. Assumption \ref{as:transition_smooth} is a continuity assumption on the transition law that is easily satisfied by most physical systems. Assumption \ref{as:transition_ergodic} makes sure that for every policy $\policy_{\Te}$, there exists a unique invariant (stationary) measure and the Markov chain reaches stationarity geometrically fast; see \cite{HLj12}. All the results hold without the transition law being geometrically ergodic. Assuming geometric ergodicity makes simulating from the ergodic measure (in Algorithm\ref{alg:SGJD}, Sec.\ref{sec:SGJD}) more meaningful.
Regarding the algorithmic conditions: Assumptions~\ref{as:policy_smooth}-\ref{as:stepsize} are standard in stochastic gradient methods; see \cite{BT00}. Assumption~\ref{as:variance_growth} says that the stochastic gradient is always bounded by the true gradient, which can grow unbounded with $\Te$. This assumption makes sure that the martingale noise in the stochastic gradient algorithm is bounded by the true gradient; see~\cite{BT00}. 

\Prop{ \label{prop:BExC}
	Under Assumption \ref{as:reward}, the expected cost $\expcst(\Te)$ in the reinforcement learning problem~(\ref{eq:MP_p}) is a bounded real-valued function, i.e,
	\Aln{
		|\expcst(\Te)| \leq \frac{M}{1 -\disc}~\forall~\Te \in \CapT.
	}
}
The following result makes sure that the stochastic gradient estimates so obtained are representative of the true gradient.

\Thmw{\label{thm:UGD}
	The stochastic gradient obtained in (\ref{eq:SGD-JD}) is an unbiased estimate of the true gradient $\nabla \expcst(\Te)$, i.e,
	\Aln{
		\EE \Blc \anab J(\Te) \Brc = \nabla \expcst(\Te).
	}
}
\Dis{
A proof is included in the Appendix. Theorem~\ref{thm:UGD} says that the estimates of the stochastic gradient are unbiased in expectation. This is required to ensure the almost sure convergence of the iterates to a stationary point~\cite{SB17}.
}

\Thmw{ \label{thm:CSG}
	Consider the sequence of policy parameters generated by Algorithm~\ref{alg:SGJD}. Under Assumptions \ref{as:reward} - \ref{as:variance_growth}, the sequence of iterates $\{\theta_k\}$ satisfies
	\Aln{
		\Te_{\tim} \rightarrow \Te^*,~\text{where}~\nabla \expcst(\Te^*) = 0,~\text{almost surely}. 
	}
}
\Dis{
A proof is included in the Appendix. The expected cost function $\expcst(\Te)$, under model assumptions, is continuous and $\lip -$ Lipschitz; see [Chapter~7]~\cite{BR11} and \cite{Hin05}. Theorem~\ref{thm:CSG} says that the sequence of iterates $\{\Te_{\tim}\}$ converges to $\Te^*$ with probability one, and since $\expcst(\Te)$ is a continuous function, $\expcst(\Te_{\tim})$ converges to $\expcst(\Te^*)$ with probability one. The gradient (which can be unbounded) at iterates $\{\Te_{\tim}\}$ is such that $\nabla \expcst(\Te^*) = 0$ with probability one.  }
\subsection{Sample Complexity}\label{subsec:complexity}
In this section, we consider the convergence rate analysis of the PG-JD algorithm. We choose the stepsize to be $\stepsize_{\tim} = \tim^{-b}$ for some parameter $b \in (0,1)$. Since the optimization of $\expcst(\Te)$ is generally non-convex,
we consider the convergence rate in terms of a metric of non-stationarity, i.e., the norm of
the gradient $\|\nabla \expcst (\Te) \|^2$. The following theorem considers a diminishing step-size and establishes a $\BigO(1/ \sqrt{k})$ rate for the decrement of the expected gradient norm square $\|\nabla \expcst (\Te_{\tim}) \|^2$.
%
\Thmw{ \label{thm:smpl_cmp}
Let $\Blc \Te_{\tim}\Brc_{\tim \geq 0}$ be the sequence of parameters of the policy $\policy_{\Te_{\tim}}$ generated by Algorithm~\ref{alg:SGJD}. Let the stepsize be $\stepsize_{\tim} = \tim^{-b}$ for $b \in (0,1)$ and $\Delta = \min \Blc \mDel, \extra\Brc$ for some $\mDel,\extra > 0$. Let 
\Aln{
\Bigtim_{\Delta} = \min \Blc \tim: \inf_{0 \leq \dtim \leq \tim} \|\nabla \expcst (\Te_{\dtim}) \|^2 \leq \Delta \Brc
}
denote the number of iteration steps for the norm of the expected cost to come within the error neighbourhood. Then,
\Aln{
\Bigtim_{\Delta} = \BigO(\Delta^{-1/ p}),~\text{where}~p = \min\Blc 1-b, b \Brc,
}
where optimizing the complexity bound over $b$, we have $b = 1/2$. Therefore, $\Bigtim_{\Delta} = \BigO(\Delta^{-2})$.
}
\Dis{
A proof is included in the Appendix. Theorem~\ref{thm:smpl_cmp} characterizes the iteration complexity, which is a measure of the number of iteration steps of the algorithm are required to settle down on a stationary point of the value function. The iteration complexity is $\BigO(1/ \sqrt{k})$ showing that the convergence rate is similar to the stochastic gradient methods for convex settings. 
}
\Corr{ \label{cor:iter_cmp}
Let $\disc$ denote the discount factor and $\Bigtim_{\Delta}$ denote the iteration complexity. The average sample complexity $\SamCmp$ using Algorithm~\ref{alg:SGJD} is given as:
\Aln{
\SamCmp = \Bln \frac{1+\disc}{1-\disc} \Brn \Bigtim_{\Delta}.
}
}
\Dis{
A proof is included in the Appendix. Corollary~\ref{cor:iter_cmp} characterizes the sample complexity, which is a measure of the number of the expected total number of actions and states realized. Higher the discount factor $\disc$, longer the two (random) Monte-Carlo roll-outs (trajectories) that need to simulated, and hence higher the sample complexity. Together the complexity results, Theorem~\ref{thm:smpl_cmp} and Corollary~\ref{cor:iter_cmp}, provide an estimate of the duration and expected number of simulations to learn a stationary solution for the reinforcement learning task considered.
}
%
\subsection{Variance Analysis}\label{subsec:variance}
In this section, we provide an analysis of the variance of the stochastic gradient estimates obtained using weak derivatives and score function approaches. Since the $Q-$function estimation in the computation of the gradient is performed using random Monte Carlo roll-outs as in \cite{Pat18}, the stochastic gradient obtained is a function of the geometric random variable $\simT$ that characterizes the roll-out (trajectory) length. To obtain a comparison of the different methods -- weak derivatives and score function -- we consider the expected variance of the gradient estimates. A proof of Theorem~\ref{thm:var} is given in the Appendix. The proof of Theorem~\ref{thm:SF} is similar and hence omitted.
\Thmw{ \label{thm:var}
The expected variance of the gradient estimates $\anab \expcst$ obtained using weak derivatives is given as:
\Aln{
\EE \Blc \var^{WD}(\anab \expcst_{\simT}(\Te)) \Brc \leq \frac{2 \cdot M^2 \cdot \varWD}{(1 - \disc)^5},
}
where $\varWD = \EE_{\state \sim \policy_{\Te}} \Blc \| g(\Te,\state) \|^2 \Brc$.
}
\Thmw{ \label{thm:SF}
The expected variance of the gradient estimates $\anab \expcst$, if score function is used instead of weak derivatives, is given as:
\Aln{
\EE \Blc \var^{SF}(\anab \expcst_{\simT}(\Te)) \Brc \leq \frac{ M^2 \cdot \varSF}{(1 - \disc)^5},
}
where $\varSF = \EE_{(\state,\action) \sim \policy_{\Te}(\action \given \state)} \Blc \| \nabla \policy_{\Te}(\action \given \state) \|^2 \Brc$.
}
\Corr{ \label{cor:gauss_var}
For the Gaussian  policy $\policy_{\Te}(\cdot \given \state) = \normal(\Te^{\tpose}\feature(\state), \sigma^2)$, we have
\Aln{
\varWD = \frac{1}{2 \cdot \pi} \varSF.
}}
Hence, the maximum expected variance of the gradient estimates using weak derivatives is smaller than those obtained using the score function method.


\begin{figure}[t]
	\centering
	\includegraphics[scale=0.45]{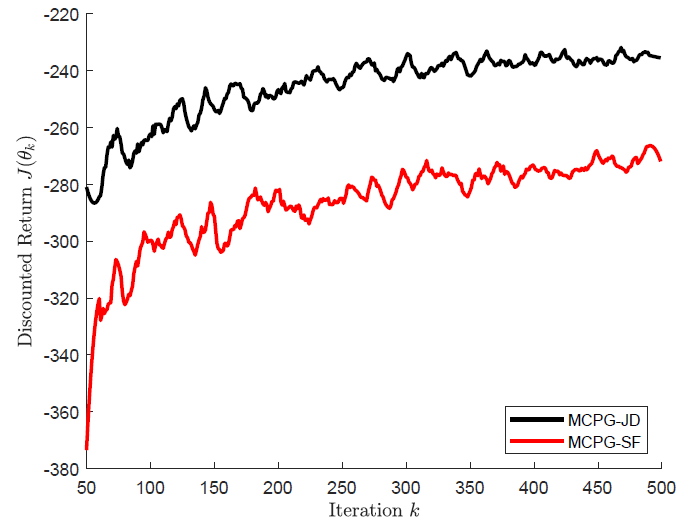}
	{
	\caption{The convergence of the discounted return as a function of the number of iterations of the policy gradient algorithms. Here at each iteration~$k$, the discounted return $J(\theta) = \mathbb{E}_{\pi_{\theta}}\{\sum_{\tim=0}^{\infty} \disc^\tim \reward(\state_{\tim},\action_{\tim})\}$ is evaluated over $50$ trajectories with $\disc = 0.97$. Observe that the discounted return is higher on average using Monte-Carlo PG-JD as opposed to PG-SF. It can be attributed to algorithm iterates converging to a ``better" stationary point due to smaller variance in the gradient estimates.}\label{fig:offset_comp}}
\end{figure}

\section{Numerical Studies} \label{sec:NumStu}
In this section, we present a simple experiment using PG-JD algorithm on the Pendulum environment in OpenAI gym~\cite{BCPSSTZ16}. The performance is compared  with Monte Carlo Policy Gradient using Score Function (PG-SF) which is akin to REINFORCE~\cite{sutton2000policy} with random roll-out horizons; see Fig.\ref{fig:offset_comp}. In the simulation environment, the pendulum starts at a random position, and the goal is to swing it up so that it stays upright. The environment state is a vector of dimension three, i.e.,  $\state_{\tim}=(\cos(\varphi_\tim),\sin(\varphi_\tim),\dot{\varphi_\tim})^\top$, where $\varphi_\tim$ is the angle between the pendulum and the upright direction, and $\dot{\varphi}_\tim$ is the derivative of $\varphi_\tim$. The action $\action_{\tim}$ is a one-dimensional scalar modified using a $\tanh$-\textit{function}, and represents the joint effort. \\
The received reward $\reward(\state_{\tim},\action_{\tim})$ is given as
\Aln{\label{equ:reward_Pendulum}
\reward(\state_{\tim},\action_{\tim}):=-(\varphi_\tim^2 + 0.1*\dot{\varphi_\tim}^2 + 0.001*{\action_{\tim}}^2),
}
which lies in $[-16.2736044,~0]$, $\varphi_\tim$ is normalized between $[-\pi,\pi]$ and $\action_{\tim}$ lies in $[-2,2]$.
The transition dynamics are determined according to Newton's Second Law of Motion. We use Gaussian policy $\pi_\theta$, which is parameterized as $\pi_\theta(\cdot\given \state)=\mathcal{N}(\theta^{\simT} \phi(\state),\sigma^2)$, where $\sigma=1.0$ and $\phi(\state) (=\state)$ being the feature vector. The policy is a stationary policy (time-homogeneous) as it is well known~\cite{Ber05} to be sufficient for infinite or random horizon discounted MDP problems. Observe that the discounted return is higher on average using PG-JD as opposed to PG-SF, which may attributable to the variance-reduced properties of the policy gradient estimates using signed measures as compared with the score function.

\Rem{It is noted that for common parametrizations of the mean of the Gaussian policy~\cite{Doy00}, for example like linear -- $\theta^{\simT} \phi(\statesim)$, the score function is unbounded with respect to $\theta$ with the expression being $\frac{(\action-\theta^{\simT} \phi(\statesim))}{\sigma^2}\phi(\statesim)$. This results in convergence issues in policy gradient algorithms for unbounded $\theta$ and unbounded state spaces. However, using Jordan decomposition, even with linear parametrization and unboundedness, the convergence of the policy gradient algorithm is ensured due to the absence of explicit function of $\theta$.}


\bibliographystyle{IEEEtran}


\appendix
\section{Proofs}
%
\subsection{Proof of Theorem~\ref{thm:PGJD} }\label{apx_A}
We begin by computing the derivative of the value function with respect to policy $\pi$, assuming that the policy is parameterized by a vector $\theta\in\mathbb{R}^d$.
\AlnN{
\nabla J(\theta)&=\int_{\state \in \statespace, \action \in \actionspace}\sum_{\tim=0}^\infty\disc^\tim \cdot \trans(\state_\tim = \state \given \state_0,\policy_\Te) \nonumber \\
&\qquad \times \nabla \policy_{\Te}(\action \given \state)\cdot \Qu_{\policy_\theta}(\state ,\action )d\action d\state  \nonumber \\
&\ =\frac{1}{1-\disc}\int_{\state \in \statespace, \action \in \actionspace}(1-\disc)\sum_{\tim=0}^\infty \disc^\tim \cdot \trans(\state_\tim = \state \given \state_0,\policy_\Te) \nonumber \\ 
& \ \qquad\times \nabla \policy_{\Te}(\action \given \state)\cdot \Qu_{\policy_\Te}(\state,\action) d\action d\state }
By Hahn-Jordan decomposition (\cite{Bil08}) for the policy gradient,
\Aln{
\nabla J(\theta)&=\frac{1}{1-\disc}\int_{\state \in \statespace, \action \in \actionspace} \ergod_{\policy_\Te}(\state)\cdot g(\Te,\state) \nonumber \\ 
 &\quad\times \{\policy^{\poT}_\Te(\action \given \state) - \policy^{\neT}_\Te (\action \given \state)\}\cdot \Qu_{\policy_\Te}(\state,\action) d\action d\state \label{equ:sgn_meas}\\
&=\frac{1}{1-\disc} \Bls \EE_{(\state,\action)\sim \policy^{\poT}_{\Te}(\cdot,\cdot)} \Blc g(\Te,\state)\cdot \Qu_{\policy_\Te}(\state,\action) \Brc \nonumber \\
 &\quad- \EE_{(\state,\action)\sim \policy^{\neT}_{\Te}(\cdot,\cdot)} \Blc g(\Te,\state) \cdot \Qu_{\policy_\Te}(\state,\action) \Brc \Brs \label{equ:policy_grad}
}
where we defime the ergodic measure \cite{SB17} $\ergod_{\policy_\Te}(\state) \defeq (1-\disc)\sum_{\tim=0}^\infty \disc^\tim \cdot \trans(\state_\tim = \state \given \state_0,\policy_\Te)$, the positive induced measure $\ergod_{\policy_\Te}(\state) \cdot \policy^{\poT}_\Te(\action \given \state) \defeq \policy^{\poT}_{\Te}(\state,\action)$ and the negative induced measure $\ergod_{\policy_\Te}(\state) \cdot \policy^{\neT}_\Te(\action \given \state) \defeq \policy^{\neT}_{\Te}(\state,\action)$.  \hfill $\blacksquare$

\subsection{Proof of Theorem~\ref{thm:UnbEst}}\label{apx_B}
Equation (\ref{eq:PaCs}) follows by definition of the path-wise cost. We will prove (\ref{eq:UnQ}) below. Here $T$ is a geometric random variable. 
\Aln{
\EE_{\simT} \Blc \hat{\Qu}_{\policy_{\Te}}(\state,\action;\simT)\Brc &= \EE_{\simT} \Blc \EE_{\policy_{\Te}}^{x} \Blc \sum_{\tim = 0}^{\simT} \reward(\state_{\tim},\action_{\tim}) \Brc \Brc  \\
&= \EE_{\simT} \Blc \EE_{\policy_{\Te}}^{x} \Blc \sum_{\tim = 0}^{\infty} \mathds{1}(\simT \geq \tim)\reward(\state_{\tim},\action_{\tim}) \Brc \Brc \label{eq:UnbEx} \nonumber}
By boundedness of rewards and using Fubini's Theorem \cite{Bog07}, (\ref{eq:UnbEx}) can be written as:
\Aln{
\EE_{\simT} \Blc \hat{\Qu}_{\policy_{\Te}}(\state,\action;\simT)\Brc 
= \EE_{\policy_{\Te}}^{x} \Blc \EE_{\simT} \Blc \sum_{\tim = 0}^{\infty} \mathds{1}(\simT \geq \tim)\reward(\state_{\tim},\action_{\tim}) \Brc \Brc \nonumber }
By Linearity of Expectation with bounded rewards,
\Aln{
\EE_{\simT} \Blc \hat{\Qu}_{\policy_{\Te}}(\state,\action;\simT)\Brc 
&= \EE_{\policy_{\Te}}^{x} \Blc  \sum_{\tim = 0}^{\infty} \EE_{\simT} \Blc \mathds{1}(\simT \geq \tim)\Brc  \reward(\state_{\tim},\action_{\tim}) \Brc \nonumber \\
&\quad= \EE_{\policy_{\Te}}^{x} \Blc  \sum_{\tim = 0}^{\infty} \probmeas (\simT \geq \tim)  \reward(\state_{\tim},\action_{\tim}) \Brc \nonumber \\
\probmeas (\simT \geq \tim) = \disc^{\tim},~ &\text{by virtue of geometric distribution.} \nonumber }
Therefore, $\EE_{\simT} \Blc \hat{\Qu}_{\policy_{\Te}}(\state,\action; \simT) \Brc =  \Qu_{\policy_{\Te}}(\state,\action)$
\hfill $\blacksquare$
%
\subsection{Proof of Theorem~\ref{thm:UGD}}\label{apxC}
Consider the stochastic gradient in (\ref{eq:SGD-JD}) when used with the two simulator policies associated with the positive and negative induced measures,
\Aln{
\anab J(\Te) &= \frac{g(\Te,\state_0)}{1-\disc} \Bls \pathcst_{\hatpol_{\Te}^{\poT}}^{\simT_{\randsamp}} - \pathcst_{\hatpol_{\Te}^{\neT}}^{\simT_{\randsamp}} \Brs.\\
\hatpol_{\Te}^{\poT} &\defeq \{\policy^{\poT}_{\Te}, \{\policy_{\Te} \}_l \}, l = 1,2,\cdots \nonumber \\
\hatpol_{\Te}^{\neT} &\defeq \{\policy^{\neT}_{\Te}, \{\policy_{\Te} \}_l \}, l = 1,2,\cdots  \nonumber 
}
Using Ionescu Tulcea theorem \cite{Nev65,HL12}, define the induced probability measures (as in (\ref{eq:ITthm})) $\inducedPMP$ and $\inducedPMN$ for some initial state $\state_0$. Let the expectation operator 
\Aln{
\EE_\statesim \defeq \inducedEMP \inducedEMN,
}
indicate the expectation with respect to measures $\inducedPMP$ and $\inducedPMN$ respectively.
By Fubini's Theorem \cite{Bog07}, result (\ref{eq:UnQ}), and by the property of the expectation $\inducedEMN \Blc \Qu_{\hatpol_{\Te}^{\poT}}(\state_0,\action_\arbindxp) \Brc = \Qu_{\hatpol_{\Te}^{\poT}}(\state_0,\action_\arbindxp)$ and $\inducedEMP \Blc \Qu_{\hatpol_{\Te}^{\neT}}(\state_0,\action_\arbindxn) \Brc = \Qu_{\hatpol_{\Te}^{\neT}}(\state_0,\action_\arbindxn),$ we have 
\Aln{ \label{eq:ind_grad}
\!\!\!\EE_\statesim \Blc \anab J(\Te)\Brc = \frac{g(\Te,\state_0)}{1-\disc} \Bls \Qu_{\hatpol_{\Te}^{\poT}}(\state_0,\action_\arbindxp) -  
\Qu_{\hatpol_{\Te}^{\neT}}(\state_0,\action_\arbindxn) \Brs
}
Let the expectation operator 
\Aln{
\EE \defeq \EE_{\state_0 \sim \ergod_{\policy_{\Te}}(\state)} \EE_\statesim,
}
indicate the expectation with respect to the ergodic measure and induced measures.
Taking $\EE_{\state_0 \sim \ergod_{\policy_{\Te}}(\state)}$ on both sides of~(\ref{eq:ind_grad}), the result follows.
\hfill $\blacksquare$
\subsection{Proof of Theorem~\ref{thm:CSG}}\label{apxD}
The proof uses arguments similar to \cite{BT00}. Below we highlight the additional results that are required. Consider the stochastic gradient algorithm for computing the optimal stationary randomized policy of the reinforcement learning problem (\ref{eq:MP_p}).
\Aln{ \label{eq:PrEq}
\Te_{\tim+1} = \Te_{\tim} + \stepsize_\tim \cdot \anab \expcst(\Te_{\tim}),~\text{for}~\tim = 0,1,\cdots.
}
The stochastic gradient expression in (\ref{eq:PrEq}) for $\tim = 0,1,\cdots$ can be rewritten as:
\Aln{ \label{eq:PrEq1}
\Te_{\tim+1} = \Te_{\tim} + \stepsize_\tim \cdot \Blc \nabla \expcst(\Te_{\tim}) - \Bls \nabla \expcst(\Te_{\tim}) - \anab \expcst(\Te_{\tim})\Brs \Brc.
}
Let the noise $\noise_{\tim} \defeq \nabla \expcst(\Te_{\tim}) - \anab \expcst(\Te_{\tim})$. The stochastic gradient algorithm can now be written as:
\Aln{ \label{eq:MnEq}
\Te_{\tim+1} = \Te_{\tim} + \stepsize_\tim \cdot \Blc \nabla \expcst(\Te_{\tim}) - \noise_{\tim} \Brc,~\text{for}~\tim = 0,1,\cdots.
}
\Prop{\label{prop:Nse}
Let $\filtr_{\tim}$ denote the sigma algebra defined by the set $\{\simT_{\tim},\{(\statesim^{\poT}_{\arbindxn},\action^{\poT}_{\arbindxn}),(\statesim^{\neT}_{\arbindxn},\action^{\neT}_{\arbindxn})\}_{\arbindxn = 0,1,\cdots, \simT_{\tim}-1} \}$. The noise $\noise_{\tim}$ has the following moments:~
\begin{itemize}
\item[i.)] The expectation $\EE \{\noise_{\tim} \given \filtr_{\tim} \} = 0$, and
\item[ii.)] The variance  $\EE \Blc \nrm \noise_{\tim} \nrm^{2} \given \filtr_{\tim} \Brc \leq \const_1 + \const_2  \nrm \nabla \expcst(\Te_{\tim})
 \nrm^{2}$ for some $\const_1,\const_2 > 0$.
\end{itemize}
}
\Prf[Proof of Proposition~\ref{prop:Nse}]{
The proposition follows easily from the Theorem~\ref{thm:UGD} and assumption (A.A3) in Sec.\ref{subse:MoDAs} on the stochastic gradient.
}
\Prop{\label{prop:SerConv}
Let $\y_{\tim}$, $\z_{\tim}$, and $\w_{\tim}$ be three sequences such that $\w_{\tim} \geq 0~\forall {\tim}$. Suppose the series $\sum_{\tim=0}^{\simT} \z_{\tim}$ converges as $\simT \rightarrow \infty$. Suppose the series 
\Aln{\label{eq:Yser}
\y_{\tim+1} \geq \y_{\tim} + \w_{\tim} - \z_{\tim}~\text{and}~|\y_{\tim}| < \infty ~\forall{\tim}.
}
Then the following holds:
\Aln{
\y_{\tim} \rightarrow \y^* (< \infty)~~\text{and}~~\sum_{\tim=0}^{\infty} \w_{\tim} < \infty. 
}
} 
\Prf[Proof of Proposition~\ref{prop:SerConv}]{
For arbitrary $\hat{\tim} > 0$, we have from (\ref{eq:Yser}) 
\Aln{
\y_{\hat{\tim}+\arbindxn} \geq \y_{\hat{\tim}} - \sum_{\tim \geq \hat{\tim}}^{\arbindxn} \z_{\tim}. \nonumber}
By letting $\hat{\tim} \rightarrow \infty$ and $\arbindxn \rightarrow \infty$, we have $\liminf_{\arbindxn \rightarrow \infty} \y_{\arbindxn} \geq \y_{\hat{\tim}} - \sum_{\tim \geq \hat{\tim}}^{\infty} \z_{\tim}$, and therefore
\Aln{ \liminf_{\arbindxn \rightarrow \infty} \y_{\arbindxn} \geq \limsup_{\hat{\tim} \rightarrow \infty}\y_{\hat{\tim}} - \lim_{\hat{\tim} 
\rightarrow \infty} \sum_{\tim \geq \hat{\tim}}^{\infty} \z_{\tim}. \label{eq:infsup}
} 
As $|\y_{\tim}| < \infty ~\forall{\tim}$, (\ref{eq:infsup}) implies that $\lim_{\tim} \y_{\tim} \rightarrow \y^*$. Also,
\Aln{
\sum_{\tim}^{\simT} \w_{\tim} \leq \y_{\tim+1} - \y_0 + \sum_{\tim = 0}^{\simT} \z_{\tim}.\nonumber }
Therefore we may conclude
 \Aln{ \lim_{\simT \rightarrow \infty} \w_{\tim}< \infty ~\text{as}~\y_{\tim}~\text{and}~ \sum_{\tim =0}^{\infty} \z_{\tim}~\text{converge}. \nonumber
}
Therefore, the two series $\{ \y_{\tim}\}$ and $\{ \w_{\tim} \}$ converge to a finite value.
}
Let $\delta >0$ be an arbitrary positive number and let $\eta(\delta)$ be a constant depending on $\delta$. As in \cite{BT00}, partition the set of all times $\tim \in \nat$ into intervals $\negFI$ and $\fullint_\tim$ such that 
\Aln{ \label{eq:defInt}
\nrm \nabla \expcst(\Te_{\tim}) \nrm \geq \delta~\forall \tim \in \fullint_{\tim},~\text{and}~\nrm \nabla \expcst(\Te_{\tim}) \nrm < \delta~\forall \tim \in \negFI.
}
\Prop{\label{prop:FI}
The expected cost $\expcst(\Te)$ increases by a fixed amount on the intervals $\fullint_{\tim}~\forall \tim \in \nat$.
}
\Prf[Proof of Proposition~\ref{prop:FI}]{
The result follows from Lemma~$5$ in \cite{BT00}.
}
\Prop{\label{prop:FMI}
There are finitely many intervals $\fullint_{\tim}$.
}
\Prf[Proof of Proposition~\ref{prop:FMI}]{
The result follows from Proposition~\ref{prop:BExC} and Proposition~\ref{prop:FI}.
}

Define the indicator function
$\indF_{\tim}=
\begin{cases}
1~~\text{if}~\tim \in \negFI \\
0~~\text{otherwise}.
\end{cases}$ \\
From Proposition~\ref{prop:FMI}, there exists a $\tim_{0}$ such that for all $\tim \geq \tim_{0}$, we have $\indF_{\tim} = 1$. Therefore, by (\ref{eq:defInt}), we have
\Aln{ \label{eq:Limsup}
\limsup_{\tim \rightarrow \infty} \nrm \nabla \expcst(\Te_{\tim}) \nrm \leq \delta.
}
By Taylor's expansion, we have for the expected cost $\expcst(\Te)$,
\Aln{ \label{eq:TEexp}
\expcst(\Te_{\tim+1}) &\geq \expcst(\Te_{\tim}) + \stepsize_{\tim} \nrm \nabla \expcst(\Te_{\tim}) \nrm^2 \nonumber \\
 &- \Blc \stepsize_{\tim} \noise_{\tim}^{\tpose} \nrm \nabla \expcst(\Te_{\tim}) \nrm + \lip \stepsize_{\tim}^2 \nrm \noise_{\tim} \nrm^2 \Brc. \nonumber \\
}
From (\ref{eq:Limsup}), for $\tim > \tim_{0}$,
\Aln{ \label{eq:TEexp2}
\!\!\!\!\expcst(\Te_{\tim+1}) &
\geq \expcst(\Te_{\tim}) - \indF_{\tim}  \Blc \stepsize_{\tim} \noise_{\tim}^{\tpose} \nrm \nabla \expcst(\Te_{\tim}) \nrm + \lip \stepsize_{\tim}^2 \nrm \noise_{\tim} \nrm^2 \Brc. 
}
From Lemma~$3$ in \cite{BT00}, the series $\indF_{\tim}  \Blc \stepsize_{\tim} \noise_{\tim}^{\tpose} \nrm \nabla \expcst(\Te_{\tim}) \nrm + \lip \stepsize_{\tim}^2 \nrm \noise_{\tim} \nrm^2 \Brc$ converges. \\ From Proposition~\ref{prop:SerConv}, with $\w_{\tim} = 0~\forall \tim$, and $\z_{\tim} = \indF_{\tim} \cdot \Blc \stepsize_{\tim} \noise_{\tim}^{\tpose} \nrm \nabla \expcst(\Te_{\tim}) \nrm + \lip \stepsize_{\tim}^2 \nrm \noise_{\tim} \nrm^2 \Brc$, the iterates $\{\expcst(\Te_{\tim}) \}$ converge to a finite value $\expcst(\Te^*)$. We conclude that $\nabla \expcst(\Te^*) = 0$ as $\delta >0$ in (\ref{eq:defInt}) was arbitrary. \hfill $\blacksquare$


\subsection{Proof of Theorem~\ref{thm:smpl_cmp}}\label{apx_E}
Begin by considering the Taylor expansion of the objective $J(\theta)$ along the line between $\theta_k$ and $\theta_{k+1}$:
\Aln{\label{eqn:begin_rate}
\expcst(\Te_{\tim+1}) \geq \expcst(\Te_{\tim}) &+ \epsilon_{\tim} \nrm \nabla \expcst(\Te_{\tim}) \nrm^2 - \epsilon_{\tim} \noise^{\simT}_{\tim} \nabla \expcst(\Te_{\tim}) \nonumber \\
 & - \epsilon^2_{\tim} \lip \nrm \noise_{\tim} \nrm^2. }
Now compute the total expectation of both sides to write
\Aln{\label{eqn:expected_taylor}
\EE \Blc \expcst(\Te_{\tim+1}) \Brc \geq \EE \Blc \expcst(\Te_{\tim}) \Brc &+ \epsilon_{\tim} \EE \Blc \nrm \nabla \expcst(\Te_{\tim}) \nrm^2 \Brc \\ 
&- \epsilon^2_{\tim} \lip \EE \Blc \nrm \noise_{\tim} \nrm^2 \Brc.\nonumber }
Recall from Proposition~\ref{prop:Nse} that we have
\AlnN{
\EE \Blc \nrm \noise_{\tim} \nrm^{2} \given \filtr_{\tim} \Brc \leq \const_1 &+ \const_2  \EE \Blc \nrm \nabla \expcst(\Te_{\tim}) \nrm^{2} \Brc. }
which we may apply to the last term on the right-hand side of~\eqref{eqn:expected_taylor} to obtain
%
%
\AlnN{
\mathbb{E}[\!J(\!\theta_{k+1}) ] \!\geq \mathbb{E}[J(\!\theta_k\!)] + \epsilon_{\tim} (\!1 \!-\! \lip \epsilon_{\tim} \const_2\!) \EE \Blc \!\nrm \nabla \expcst(\!\Te_{\tim})\! \nrm^{2} \!\! \Brc - \epsilon^2_{\tim} \lip \const_1. }
after gathering like terms. 
Let $\tim_0$ be such that $(1 - \lip \epsilon_{\tim_0} \const_2) \approx 1$. If $0 <\extra << 1$ is such that$\lip \epsilon_{\tim_0} \const_2 = \extra$, then $\tim_{0} = (\const_2 \lip / \extra)^{1/b}$. For all $\tim > \tim_0$, we have
\AlnN{
\EE \Blc \nrm \nabla \expcst(\Te_{\tim}) \nrm^{2} \Brc &\leq \frac{1}{\epsilon_{\tim}} \Blc \mathbb{E}[J(\theta_{k+1})] -  \mathbb{E}[J(\theta_k)] \Brc + \epsilon_{\tim} \lip \const_1. }
Let the shifted time scale $\shifttim = \tim - \tim_0$. For $\shifttim = 0,1,2,\cdots$
\AlnN{
\EE \Blc \nrm \nabla \expcst(\Te_{\shifttim}) \nrm^{2} \Brc &\leq \frac{1}{\epsilon_{\shifttim}} \Blc  \mathbb{E}[J(\theta_{\shifttim+1})] -  \mathbb{E}[J(\theta_{\shifttim})]  \Brc + \epsilon_{\shifttim} \lip \const_1.}
Summing $\shifttim = 1$ to $\Ulim$ terms, we have
\AlnN{
\! \sum_{\shifttim = 1}^{\Ulim}\! \EE \Blc &\nrm \nabla \expcst(\Te_{\shifttim}) \nrm^{2} \Brc \leq \sum_{\shifttim = 1}^{\Ulim}\!\! \Bln \frac{1}{\!\epsilon_{\shifttim}} \!-\! \frac{1}{\epsilon_{\shifttim - 1}}\! \Brn    \mathbb{E}[J(\theta_{\shifttim}\!)]  \\ 
&\quad+ \frac{1}{\epsilon_{\Ulim}} \mathbb{E}[\expcst(\Te_{\Ulim+1}) ]+ \frac{1}{\epsilon_{0}}  \mathbb{E}[\expcst(\Te_{0})]
 + \sum_{\shifttim = 1}^{\Ulim} \epsilon_{\shifttim} \lip \const_1. }
Now upper-estimate the right-hand side of the preceding expression by its absolute value, applying Jensen's inequality as $|\mathbb{E}[\expcst(\Te)]|\leq \mathbb{E}[|\expcst(\Te) |]$ to obtain
\AlnN{
\sum_{\shifttim = 1}^{\Ulim} \EE \Blc \nrm \nabla \expcst(\Te_{\shifttim}) \nrm^{2} \Brc &\leq \sum_{\shifttim = 1}^{\Ulim} \Bln \frac{1}{\epsilon_{\shifttim}} - \frac{1}{\epsilon_{\shifttim - 1}} \Brn   \mathbb{E}[ |\expcst(\Te_{\shifttim})|]  \\ 
&\quad+ \frac{1}{\epsilon_{\Ulim}}  \mathbb{E}[ | \expcst(\Te_{\Ulim+1})|]  + \frac{1}{\epsilon_{0}}  \mathbb{E}[| \expcst(\Te_{0})|] \\ 
&\quad+ \sum_{\shifttim = 1}^{\Ulim} \epsilon_{\shifttim} \lip \const_1.}
Applying $\Bigabs \expcst(\Te_{\shifttim}) \Bigabs \leq \frac{\Upbound}{1-\disc}~\forall~\shifttim$ to the preceding expression allows us to write
\AlnN{
\sum_{\shifttim = 1}^{\Ulim}\! \EE\! \Blc \! \nrm \nabla \expcst(\Te_{\shifttim}) \nrm^{2}\! \Brc \leq
 \frac{2 \Upbound}{(1-\disc)\epsilon_{\Ulim}} + \lip \const_1  + \frac{\lip \const_1}{1-b} (\Ulim^{1-b}-1).}

Now, set $\epsilon_{\Ulim} = \Ulim^{-b}$ so that we have
\AlnN{
\frac{1}{\Ulim} \sum_{\shifttim = 1}^{\Ulim} \EE \Blc \nrm \nabla \expcst(\Te_{\shifttim}) \nrm^{2} \Brc \leq \frac{2 \Upbound}{(1-\disc)\epsilon_{\Ulim}} \Ulim^{b-1} + \frac{\lip \const_1}{1-b} \Ulim^{-b} \\
 \leq \cst_1 \Ulim^{b-1} + \cst_2 \Ulim^{-b} \\
 \leq \BigO(\Ulim^{-p}), \\ ~\text{where}~ p = \argmin \Blc 1-b, b \Brc = 1/2.}
By definition of $\Bigtim_{\mDel}$, we have 
\AlnN{
\EE \Blc \nrm \nabla \expcst(\Te_{\shifttim}) \nrm^{2} \Brc &\geq \mDel,~~\text{for}~\shifttim < \Bigtim_{\mDel}\; , 
}
Therefore, we may write
\AlnN{ \mDel \leq \frac{1}{\Bigtim_{\mDel}} \sum_{\shifttim = 1}^{\Bigtim_{\mDel}} \EE \Blc \nrm \nabla \expcst(\Te_{\shifttim}) \nrm^{2} \Brc &\leq \BigO(\Bigtim^{-p}_{\mDel}). \\
}
This implies that $\Bigtim_{\mDel} \leq \BigO(\mDel^{-1/p})$. Since $\shifttim = \Bigtim_{\mDel} = \tim - \tim_0$, we have $\tim = \Bigtim_{\mDel} + \tim_0$. But we know that $\tim_0 = \BigO(\extra^{-1/b})$. Therefore, with $\Delta = \min \Blc \mDel, \extra \Brc$, we have $\Bigtim_{\Delta} = \BigO(\Delta^{-2})$.\hfill $\blacksquare$

\subsection{Proof of Corollary~\ref{cor:iter_cmp}}\label{apx_E}
At each iteration step $\tim$, two trajectories (Monte Carlo roll-outs) of length $\simT$ are simulated. For an iteration complexity of $\Bigtim_{\Delta}$, the sample complexity can derived by considering the simulation process as a discrete time queue with geometric inter-arrival times of length $2 \simT$. Clearly, the parameter of the geometric distribution is $\frac{1-\disc}{2}$.  The distribution of the sample complexity is given by the Pascal distribution with parameters $\Bigtim_{\Delta}$ and $\frac{1-\disc}{2}$. The average sample complexity is given by the mean of the Pascal distribution, and is given as $\SamCmp = \Bln \frac{1+\disc}{1-\disc} \Brn \Bigtim_{\Delta}.$
\hfill $\blacksquare$
\subsection{Proof of Theorem~\ref{thm:var}}\label{apx_F}
Consider the gradient estimate generated by Algorithm~\ref{alg:SGJD}:
\AlnN{
\anab J_{\simT}(\Te) &= \frac{g(\Te,\state_0)}{1-\disc} \Bls \pathcst_{\hatpol_{\Te}^{\poT}}^{\simT} - \pathcst_{\hatpol_{\Te}^{\neT}}^{\simT}  \Brs. \\
\text{As $\pathcst_{\hatpol_{\Te}^{\poT}}^{\simT}$ and $\pathcst_{\hatpol_{\Te}^{\neT}}^{\simT}$}~&\text{are positively correlated, we have}\\
\var(\anab J_{\simT}(\Te)) &\leq \frac{\nrm g(\Te,\state_0) \nrm^{2}}{(1-\disc)^2} \Blc \var(\pathcst_{\hatpol_{\Te}^{\poT}}^{\simT}) + \var(\pathcst_{\hatpol_{\Te}^{\neT}}^{\simT}) \Brc. }
By Assumption \ref{as:reward} and the def. of path-wise cost $\pathcst_{\hatpol_{\Te}^{(\cdot)}}^{\simT}$,
\AlnN{\var(\anab J_{\simT}(\Te)) & \leq  \frac{\nrm g(\Te,\state_0) \nrm^{2}}{(1-\disc)^2}  \Blc \frac{2 \Upbound^2 \simT}{(1-\disc)^2} \Brc. }
Taking expectation w.r.t the geometric random variable $\simT$ and distribution over initial states, we have
\AlnN{%
\EE \Blc \var(\anab J_{\simT}(\Te)) \Brc & \leq \frac{ \EE_{\state \sim \ergod_{\policy_{\Te}}(\state)} \Blc \nrm g(\Te,\state) \nrm^{2} \Brc}{(1-\disc)^4} 2 \Upbound^2 \EE_{\simT} \Blc \simT \Brc.  }

Since $T$ is geometrically distributed with parameter $1-\gamma$, we may substitute  $\EE_{\simT} \Blc \simT \Brc = \frac{1}{1-\gamma}$ into the preceding expression. Doing so yields:
\AlnN{ \EE \Blc \var(\anab J_{\simT}(\Te)) \Brc & \leq \frac{2 \Upbound^2 \cdot \EE_{\state \sim \ergod_{\policy_{\Te}}(\state)} \Blc \nrm g(\Te,\state) \nrm^{2} \Brc}{(1-\disc)^5}. \nonumber
}\hfill $\blacksquare$
%


\newpage

\end{document}